# PREMA: Part-based REcurrent Multi-view Aggregation Network for 3D Shape Retrieval

*Jiongchao Jin\**
School of Computer Science, Beihang University
Beijing, China
jinjiongchao@buaa.edu.cn

*Huanqiang Xu*
School of Computer Science, Beihang University
Beijing, China
xuhuanqiang@buaa.edu.cn

*Pengliang Ji, Zehao Tang*
School of Computer Science, Beihang University
Beijing, China
jpl1723@buaa.edu.cn, t525792782@gmail.com

*Zhang Xiong*
School of Computer Science, Beihang University
Beijing, China
xiongz@buaa.edu.cn

*Abstract*—We propose the Part-based Recurrent Multi-view Aggregation network(PREMA) to eliminate the detrimental effects of the practical view defects, such as insufficient view numbers, occlusions or background clutters, and also enhance the discriminative ability of shape representations. Inspired by the fact that human recognize an object mainly by its discriminant parts, we define the multi-view coherent part(MCP), a discriminant part reoccurring in different views. Our PREMA can reliably locate and effectively utilize MCPs to build robust shape representations. Comprehensively, we design a novel Regional Attention Unit(RAU) in PREMA to compute the confidence map for each view, and extract MCPs by applying those maps to view features. PREMA accentuates MCPs via correlating features of different views, and aggregates the part-aware features for shape representation.

Finally, we show extensive evaluations to demonstrate that our method achieves the state-of-the-art accuracy for 3D shape retrieval on ModelNet-40 and ShapeNetCore-55 datasets.

## I. INTRODUCTION

With the rapid development of 3D sensing techniques and the proliferation of large-scale 3d shape repositories[29][5], 3D shape retrieval has become an essential topic of scene understanding. And it closely connects to the applications in robotic vision and augmented reality. A discriminative and robust 3D shape descriptor has been a long-standing pursue in the vision communities. Recently, extensive research efforts have been devoted on 3D shape representation learning based on deep neural networks.

Among the various existing methods on data-driven feature learning for 3D shapes, deep learning over multi-view projections of 3D shapes achieves the most promising performance of shape discrimination[26][17][6], thanks to rich information captured by multiple observation views and the great breakthrough of image feature learning by convolutional neural networks(CNN)[19]. Despite the high discriminative power of multi-view CNN(MVCNN), such models are typically trained with clean images rendered in a fixed, ordered set views. That extremely hinders model's utility in real world scenarios such as multi-view active recognition. In the real world scenarios, only limited number of views can be captured in a random order, due to inaccessibility caused by physical constraint. Most views are likely to be contaminated by occlusion and/or background clutter. These practical defects dramatically degrade the discriminative ability of MVCNN.

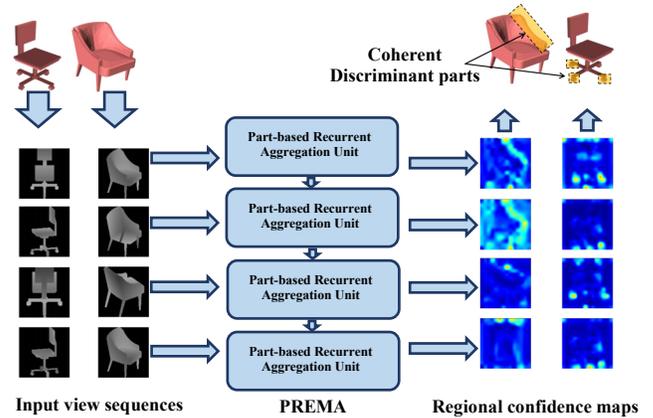

Figure 1. PREMA is proposed to reliably attend and effectively utilize coherent discriminant parts from multi-view observations. The regional confidence maps (right) generated along the sequence accentuate the discriminant parts of 3D shapes (top-right).

Targeting these challenges, our key observation is that human recognize an object by focusing on its discriminant parts (e.g. distinguishing a swivel chair from other kinds of chairs by looking at its leg part), but not necessarily the entire shape. This greatly saves the amount of information needed for recognition. Therefore, we seek for a multi-view 3D shape representation that reliably locates and effectively utilized such discriminant parts from multi-view observations. Attending to discriminant parts for more accurate visual recognition has been explored in a few previous works. However, existing methods either work with single-view images[30] or treat each view independently[34]. Our approach exploits the geometric correlation across multiple views of a 3D shape to accentuate discriminant parts.

We raise the concept of multi-view coherent parts, which is discriminant part of the object of interest that reoccurs across multiple views. Such parts are not only discriminant but also reliably trackable along a view sequence. They can be used to learn an enhanced multi-view representation for 3D shapes. To this end, we present a Part-based REcurrent Multi-view Aggregation(PREMA) Network, to accentuate discriminant parts via relating the shape information of a part across different views as shown in Figure 1. This amounts to a sequential modeling of part-based recognition using recurrent neural networks. The network can be trained in a weakly supervised manner, without the need of fine-grained part labeling. Discriminant parts are extracted and aggregated coherently along the view sequence, based on the Long-Short term Memory(LSTM) units[14].

PREMA consists of two level of LSTM units. The first level is trained to select discriminant parts in each view through gathering the coherent shape information across the view sequence. It produces a confidence map for each view feature with a Regional Attention Unit(RAU). The second level learns to aggregate the coherent discriminant part information, to form a compact 3D shape representation. In short, the per-view shape information of different parts flows in with the image sequence, so that coherent part features can be extracted and propagated along the LSTM nodes(Figure 2. ).

Extensive evaluations demonstrate that the aggregated features of coherent discriminant part possess the following advantages. First, the shape representation is highly discriminative, suited for 3D shape retrieval. Second, it is robust to missing views, object occlusion and background clutter. Also we include the comprehensive ablation studies to prove the effectiveness of our design. The main contributions of our work include:

● A recurrent aggregation network designed to learn a discriminative 3D shape representation based on multi-view coherent parts

● A regional attention unit to capture the coherent parts across different views through correlating multi-view shape information with LSTM

● Evaluations demonstrating the discriminative power in 3D shape retrieval and the robustness to view missing, object occlusion and background clutter.

## II. RELATED WORK

**View-based deep learning models.** Due to the success of deep learning in a large number of vision tasks, discriminative 3D deep representations have been extended to play an important role in view-based 3D recognition. Bai et al.[1] presented a real-time and unified 3D recognition framework, where single view feature is firstly extracted by a pre-trained CNN and an off-line re-ranking technique is proposed to merge complementary information of different level CNN features. Su et al.[26] proposed a multi-view convolutional network (MVCNN), which used a set of CNN to extract each view's deep representation and then aggregated multiple view features with the element-wise maximum operation.

Following that, Johns et al.[16] overcome the limit on the fixed-length view sequence in multi-view 3D shape identification by decomposing a view sequence into a set of view pairs. This method trains a CNN to obtain the discriminative Next-Best-View prediction, building the pairwise relationship between different view pairs. Feng et al.[8] propose a view-group-shape framework (GVCNN) to explore the content correlation among views for efficiently feature aggregation. Kanezaki et al.[17] develop a method to train CNN to learn the corresponding relationship between views and view points in a weakly supervised manner, which can urge CNN to learning more spatial information. He et al.[12] apply N-Grams algorithm to process view sequence. This method splits view sequence to overlapped unit, fuses view features in a unit and then fuses unit features to a shape feature. Chen et al.[6] focus on the problems of single view incompleteness and mult-view redundancy. They propose view-agnostic attention and view-specific attention to extract attentive view features, and then use multi-granularity view aggregation to get hierarchical shape features.

**Feature Aggregation via LSTM.** After the propose of MVCNN, the researches on 3D shape retrieval are mainly focus on how to aggregate view features into a discriminative shape feature. While the works mentioned above proposed several different aggregation methods, LSTM is more and more popular for this task in recent researches.

To exploit the correlative information from view sequence, Ma et al.[23] design a pipeline where CNNs are first used to extract view features and LSTM are employed to aggregate these features into a shape descriptor. Han et al.[9] use an encoder-RNN to process view features and a decoder-RNN to perform sequential label prediction. Attention mechanism is applied to guide which views should be paid more attention. Xu et al. [32][31] utilize extra supervision of 3D shape generation to emphasize 3D properties of shapes. They use a LSTM to fuse the information in different view features and a following max-pooling to extract a shape feature. Jiang et al.[15] split view sequence into 3 loops which are captured along 3 different directions. Considering the strong ordering in each loop, a LSTM is used to extract loop features. Then the loop features are concatenated into the final shape feature. [22] divide views into several groups based on their visual appearance and obtained raw group-level representations by the weighted sum of the view-level descriptors. LSTM is employed to exploit the context among groups to generate group-wise context features.

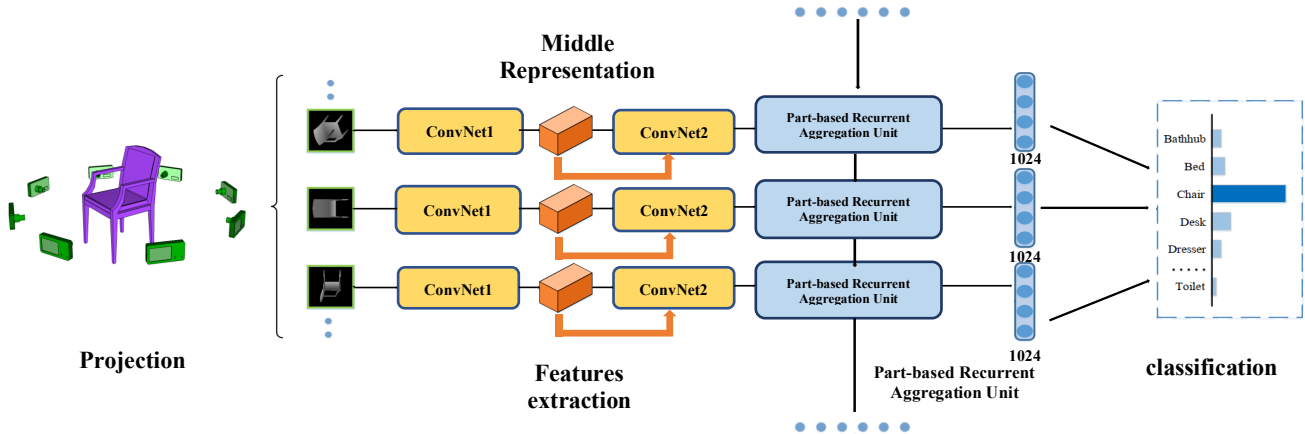

Figure 2. An overview of the architecture of Part-based REcurren

**Attention mechanism in Computer vision.** Wang *et al.*[28] develop a sophisticated self-attention mechanism to capture long-range dependency in view sequence. The proposed non-local block can be inserted into backbone network and improve the performance for several tasks. Cao *et al.*[4] design global context(GC) block to simplify Non-local block and alleviate the computational burden.

Here we stress the difference between our proposed Regional Attention Unit and their attention mechanism. In Non-local block and GC block, attention map is calculated according to correlation of pixels in one view or multi views. While our attention mechanism is built between global feature and each view feature. The global feature is delicate fusion of view features instead of raw pixels. In other words, our attention mechanism works in higher dimension.

## III. METHOD

The goal of PREMA is to produce a discriminative 3D shape representation via accentuating coherent discriminant parts across different views. Figure 2. gives the architecture of PREMA, which contains two main components: view feature extraction and part-based recurrent feature aggregation.

### A. View Feature Extraction

Given a 3D shape $x_p$ and its view sequence $S(x_p) = (p_1, p_2, ..., p_n)$, where $p_i$ denotes the projected image assigned to the *i*-th view of shape $x_p$ and n is the length of the view sequence. We use a set of CNNs to extract different view features and each view inputs the CNN independently, as shown in Figure 2. For each view $p_i$, both its middle representation $m_i$ containing abundant part information, and its final compact feature vector $v_i$ containing global shape information, are extracted for the subsequent part-based recurrent feature aggregation. Since the quality of each view feature is important to the final aggregated 3D representation, we train CNN for each view and share the same weights to make all view feature as discriminative as possible. In addition, we adopt the ResNet[11] as the backbone of our network, which has great power to learn the discriminative visual representation and fast convergence speed.

### B. Part-based Recurrent Feature Aggregation

Due to the great power of recurrent neural networks in spatial-temporal sequence, we present a part-based recurrent aggregation network to select discriminant parts in each view and coherently aggregate them alone the view sequence. Moreover, the part-based recurrent network is trained to adaptively capture the coherent, discriminant parts of images with the generated attention maps in a weakly supervised manner.

The structure of part-based recurrent aggregation is shown in Figure 3. In the process of recurrent feature aggregation, view information is propagated from the first part-based recurrent aggregation unit to the last one. At each time stamp *t*, which denotes the order of views in a sequence, the part-based recurrent aggregation unit *t* maintains the information of the recurrent unit *t* - 1 and uses it to output the current fused feature $d_t$ with the input feature $v_t$. Because the discriminative information may appear anywhere in the different views of a sequence, we adopt an element-wise max-pooling layer to combine the output features of different recurrent aggregation units, $d_1, d_2, ..., d_n$, which allows for the aggregation of information across all time steps and avoids the final representation of the view sequence biasing towards later time-steps. The final representation of view sequence $D_p$ of the 3D shape $x_p$ can be expressed as

$$D_p^i = max\,(\{d_{1,i}, d_{2,i}, ..., d_{n,i}\}) \qquad (1)$$

where $D_p^i$ denotes the *i*-th element of the feature vector $D_p$ and $d_{t,i}$ is the $i^{th}$ element of the view feature $d_i$.

**Part-based recurrent aggregation unit.** The part-based recurrent aggregation unit consists of LSTM and Regional Attention Unit (RAU), which is shown in Figure 3. It has two-level LSTMs to fuse shape information of different features. A LSTM node includes three gates: the input gate **i**, the forget gate **f** and the output gate **o**. At time stamp $t$, given input $v_t$ and previous LSTM node state $h_{t-1}$, the LSTM's update mechanism is as follows

$$i_t = \sigma(W_i v_t + U_i h_{t-1} + H_i c_{t-1} + b_i) \qquad (2)$$

$$f_t = \sigma(W_f v_t + U_f h_{t-1} + H_f c_{t-1} + b_f) \quad (3)$$
$$c_t = f_t \otimes c_{t-1} + i_t \otimes tanh(W_c v_t + U_c h_{t-1} + b_c) \quad (4)$$
$$o_t = \sigma(W_o v_t + U_o h_{t-1} + H_o c_t + b_o) \quad (5)$$
$$h_t = o_t \otimes tanh(c_t) \quad (6)$$

where $\sigma$ is the sigmoid function and $\otimes$ the vector elementwise product operator. $W_*$, $U_*$ and $H_*$ denote the weight parameters and $b_*$ is the bias vector.

The first-level recurrent layer LSTM1 is to learn the global shape information across different view representations. Specifically, at time stamp t, the LSTM1 node produces the feature vector $o_t^{(1)}$ as the aggregation of view sequence information with previous hidden state $h_{t-1}^{(1)}$ and the input global view feature $v_t$. The output feature $o_t^{(1)}$ is supposed to accumulate the discrimination information among the input view sequence before time stamp $t$, and it performs consistent discrimination for previous views. Therefore, we leverage the LSTM output feature to attend to the coherent discriminant part of 3D shape via the regional attention unit.

The regional attention unit is designed to capture long range dependencies between consistently discriminative LSTM1 output feature $o_t^{(1)}$ and the discriminant part information of middle view representation $m_t$. We adapt the computationally efficient non-local model of to accentuate the coherent discriminant parts of multiple views. As shown in Figure 3. the LSTM1 output feature $o_t^{(1)} \in \mathbb{R}^{C_o \times 1}$ and the middle representation $m_t \in \mathbb{R}^{C_m \times N}$ are first transformed into two feature spaces $h, r$ and feature dimension after transformation is $d_k$. Then they are utilized to calculate the attention, which can be expressed as

$$conf_i = \frac{exp(s_i)}{\sum_{i=1}^{N} exp(s_i)}, where, s_i = \frac{h(o_t^{(1)})^T \cdot r(m_t^i)}{\sqrt{d_k}} \quad (7)$$

where $conf_i$ indicates the confidence score for the $i$-th location of middle representation $m_t$. Note that $h(o_t^{(1)}) = W_h o_t^{(1)}, r(m_t^i) = W_r m_t^i$, where $m_i$ denotes the feature vector of the $i$-th location of middle representation $m_t$. And $W_h \in \mathbb{R}^{C_o \times d_k}$, $W_r \in \mathbb{R}^{C_o \times d_k}$ are the learned parameters, which are implemented as 1*F1 convolution or fully connected layer. Then the output of the regional attention unit is attentive part feature $attPart_t \in \mathbb{R}^{d_k \times 1}$, which can be computed as

$$attPart_t = \sum_{i=1}^{N} conf_i \cdot g(m_t^i) \quad (8)$$

where $g(m_t^i) = W_g \cdot m_t^i$ and $W_g \in \mathbb{R}^{C_m \times d_k}$ are the learned parameters, which are implemented as $1 \times 1$ convolution.

Finally, attentive part features $attPart_t$ and global LSTM1 output feature $o_t^{(1)}$ are concatenated and input to the second-level LSTM2, which aggregates the coherent discriminant part information in $attPart_t$, to produce the output feature $o_t^{(2)}$ with previous hidden state $h_{t-1}^{(2)}$ input vector $x_t$. We use the $o_t^{(2)}$ as the final aggregated feature of such part-based recurrent aggregation unit.

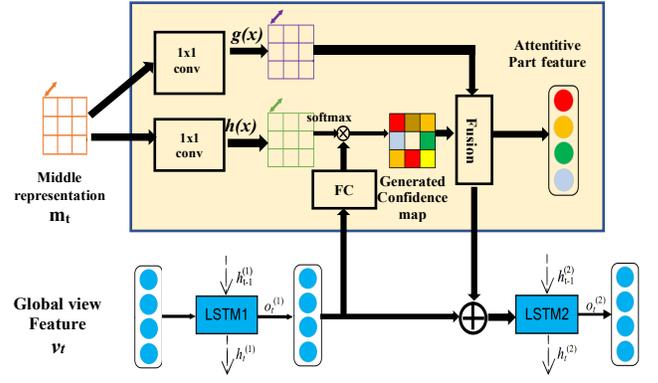

Figure 3. The structure of part-based recurrent aggregation unit.

## IV. RESULTS AND EVALUATIONS

**Implementation and experimental setting.** We use the blender script provided by [27] to render 3D shapes into multi-view images. The blender script will place 12 virtual cameras around the shape, and each one will render a $224 \times 224$ shaded image. In our experiment, we use ResNet-18 and ResNet-50[11] as the backbone network. We use the output from the third stage of ResNet as middle features and the output from the last stage as view features. In part-based recurrent aggregation unit, we use bidirectional LSTMs[25] to fuse the views features into a 2048-dimensional vector, which can be used in classification and retrieval. Note that we train the network only using cross entropy loss but no metric learning method.

We adopt a two-stage training strategy. In the first stage, we treat the view sequence as 12 irrelevant views to train a backbone network for 20 epochs. In the second stage, we load its parameters into our PREMA network and then train the network for 30 epochs. We observe the two-stage training strategy can make training smoother and obtain a better performance. In our experiment, we use a SGD optimizer for both training stages. The initial learning rate for the first training stage is 0.01 and is annealed by 0.1 at the 10th epoch. In the second stage, the initial learning rate is set to 0.001 and is anneal by 0.1 at the 20th epoch. All experiments are conducted on a server with eight Nvidia GTX 1080ti GPUs, an Intel Xeon E5 CPU and 128G RAM. We implement our method using Pytorch[24] and *all source code will be made publicly available*.

### A. Result on ModelNet40

**Datasets.** The ModelNet40[29] dataset contains 9,843 models for training, and 2,468 models for test. Most works on 3D shape retrieval only use 80/20 objects per class, while some recent works on 3D shape classification use the full dataset. For fair comparison, we follow the training/test split used in [6], using the full dataset for classification and 80/20 split for retrieval.

**Results for retrieval.** We compare PREMA with state-of-the-art methods, such as MVCNN, TCL[13], NCENet[33], VNN[12] and HEAR[6] on two metrics, mAP and AUC(Area Under Curve) of precision and recall curve, as commonly used. As is shown in TABLE I. , in retrieval tasks of ModelNet40, PREMA can achieve a better result on both AUC and mAP.

Besides, we also show our retrieval ability qualitatively in Figure 4.

**Results for classification.** We provide average accuracy over all instances and that over all classes. We compare PREMA with LFD[25](hand-crafted method), Voxception-ResNet[3](Voxel-based method), SO-Net[21](point-based method) and other view-based methods, such as MVCNN[27], Pairwise Network[16], MHBN[37], HGNN[7] and RelationNet[36]. As shown in TABLE II. , among the methods taking 12 views as input, PREMA achieves state-of-the-art performance. RotationNet[17] gets the highest points over all methods, but its way to render views from 3D shapes is quite different from others. We conduct experiments on the 20-view dataset provided by RotationNet, and achieve the same accuracy.

TABLE I. THE COMPARISON WITH STATE-OF-THE-ART METHOD ON MODELNET40. RES-18 AND RES-50 REPRESENT USING RESNET-18 AND RESNET50 BACKBONE NETWORK, RESPECTIVELY

| *Methods* | *AUC* | *mAP* |
|---|---|---|
| 3D ShapeNet[29] | 49.9 | 49.2 |
| MVCNN[26] | - | 80.2 |
| GIFT[1] | 83.1 | 81.9 |
| RED[2] | 87 | 86.3 |
| GVCNN[8] | - | 85.7 |
| TCL[13] | 89 | 88 |
| SeqViews[10] | - | 89.1 |
| VDN[20] | 87.6 | 86.6 |
| Batch-wise[35] | - | 83.8 |
| NCENet[33] | 88 | 87.1 |
| VNN[12] | 90.2 | 89.3 |
| HEAR[6] | 92.8 | 92 |
| $Ours^{Res-18}$ | 93.9 | 91.4 |
| $Ours^{Re-s50}$ | **94.6** | **92.2** |

TABLE II. 3D SHAPE CLASSIFICATION PERFORMANCES ON MODELNET40. MOST MULTI-VIEW METHODS ADOPT 12-VIEW SEQUENCE, WHILE ROTATIONNET USES 20-VIEW SEQUENCE.

| *Methods* | *Input Modality* | *Per Instance* | *Per class* |
|---|---|---|---|
| LFD | Hand-crafted | - | 75.5 |
| Vox-ResNet | Volume | 91.3 | - |
| PointNet++ | Points | 91.9 | - |
| SO-Net | Points | 93.4 | 90.8 |
| DensePoint | Points | 93.2 | - |
| MVCNN | 12-Views | 92.1 | 89.9 |
| Pairwise Net | 12-Views | - | 91.1 |
| GVCNN | 12-Views | 93.1 | - |
| MHBN | 12-Views | 94.1 | 92.2 |
| HGNN | 12-Views | **96.7** | - |
| RelationNet | 12-Views | 94.3 | 92.3 |
| HEAR | 12-Views | **96.7** | **95.2** |
| $Ours^{Res18}$ | 12-Views | 95.9 | 93.8 |
| $Ours^{Res50}$ | 12-Views | 96.1 | 94.7 |
| RotationNet | 20-Views | 97.3 | - |
| $Ours^{Res50}$ | 20-Views | **97.3** | **96.64** |

### B. Result on ShapeNetCore55

**Datasets.** The ShapeNetCore55[5] contains 51,190 models categorized into 55 categories and 204 subcategories. We adopt the official training and testing split: 70\% were used for training, 10\% for validation and 20\% for testing. This dataset has two variants: ShapeNet55 normal dataset where all shapes are consistently aligned, and ShapeNet55 perturbed dataset where the shapes are randomly oriented. We conduct our experiment on the normal one.

**Result for retrieval.** For this dataset, we take F1 score, mean Average Precision(mAP) and normalized discounted cumulative gain(NDCG) as metrics. Considering the imbalance problem of ShapeNetCore55, all metrics are computed by unweighted average over all test samples. The evaluation results of MVCNN, GIFT[1], Kd-Network[18] and VNN are available. And the results are listed in TABLE III. . While VNN gets better F1 score, PREMA achieves the best mAP score and NDCG score in all methods.

TABLE III. COMPARING RETRIEVAL PERFORMANCE ON SHAPENETCORE55.

| *Methods* | *F1* | *mAP* | *NDCG* |
|---|---|---|---|
| MVCNN | 76.4 | 87.3 | 89.9 |
| GIFT | 68.9 | 82.5 | 89.6 |
| Kd-network | 74.3 | 85.0 | 90.5 |
| VNN | **78.9** | 90.3 | 92.8 |
| $Ours^{Re-s50}$ | 62.8 | **90.8** | **93.1** |

### C. Ablation Study

In this section, we prove the effectiveness of regional attention unit empirically and investigate the impact of feature aggregation. All experiments keep the same setting with retrieval experiment on ModelNet40.

**Impact of Regional Attention Unit.** We remove the regional attention unit from PREMA and get a simpler pipeline, named doubleLSTMs, where the view features are aggregated by two bidirectional LSTMs. In our experiments, the results are shown in 0 It is clear that our proposed regional attention unit can perform better than the widely-used idea of using RNN to fuse multi-view features. The comparison is performed in 0

**Impact of feature aggregation.** For feature aggregation, we compare two fusion schemes, max pooling and LSTM layers. We also compare the performance of regular LSTM and our proposed PREMA(bidirectional LSTM). The AUC and mAP of max pooling are 91.5 and 88.2 respectively. Using LSTM can have a better result with AUC 94.0 and mAP 91.4. Our proposed PREMA(bidirectional LSTM) can achieve the best result and get 0.6 higher than LSTM on AUC and 0.8 higher on mAP. We realize that simple max pooling fails to utilize all information among views and completely ignores the information represented by the order of view sequence. Unlike max pooling, LSTM doesn't discard any features and has ability to analysis sequential features. So we argument that LSTM layers is more suitable for aggregation of view features. In addition, the view sequence is rendered from a circle, so there is no difference between the first position and the last position in the sequence. Bidirectional LSTM has advantage of understanding this point over regular LSTM, for what bidirectional LSTM achieves better performance.

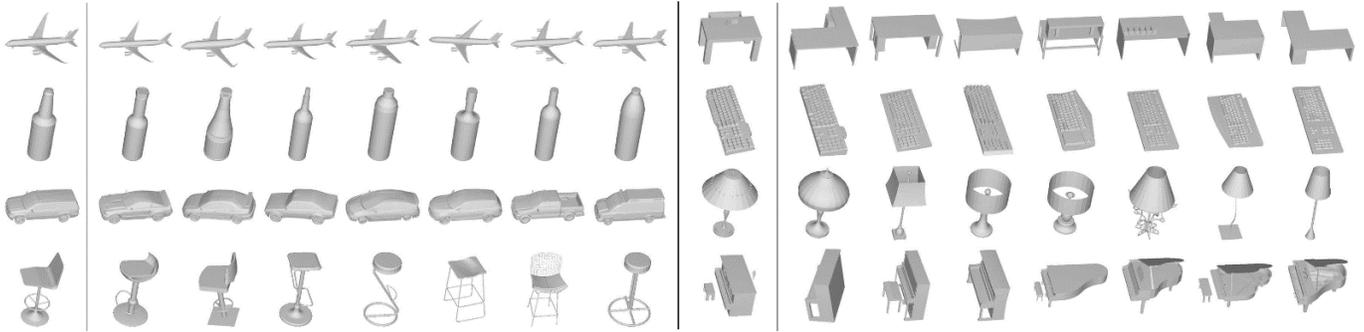

Figure 4. Top 7 retrieval result on ModelNet40 dataset. The rightmost shape in each row is the query and the rest shapes are the top 7 retrieved shapes.

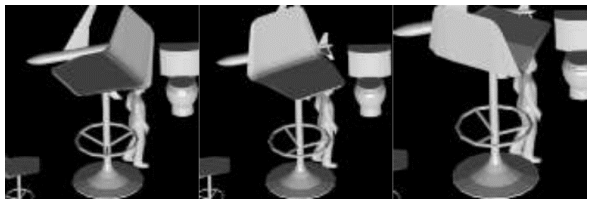

(a) background cluttered sample

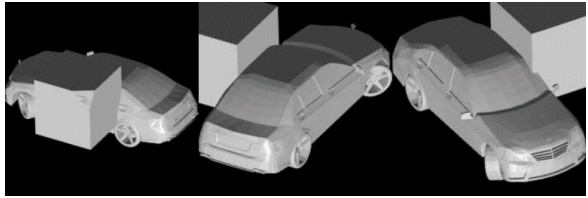

(b) object occluded sample

Figure 5. Noisy data used in robustness evaluation. (a) Views of a chair with a person, a airplane, a toilet and another chair in the background. (b) Views of a car occluded by a cube with the scale 0.8.

## D. Robustness Evaluation

We evaluate the robustness of PREMA against missing views, occlusion and background clutter. We compare with two baselines, MVCNN and doubleLSTMs. All the methods are trained on clean data, but tested on noise data, which needs better ability to capture the discriminative part of 3D shapes. More details can be found in supplementary material.

TABLE IV. THE PERFORMANCE WITH AND WITHOUT REGIONAL ATTENTION UNIT.

| Methods | AUC | mAP |
|---|---|---|
| double LSTMs | 93.6 | 90.9 |
| PREMA | 94.6 | 92.2 |

**View missing.** We evaluate the discriminative ability of our part-based recurrent aggregated features for view missing on the ModelNet40 dataset. For the testing view sequence, some images in the sequence are replaced by a black image. The length of the full sequence is 12, and we test PREMA when 2,4,6 and 8 views are missed. The performances of our method with different missing views are presented in 0

When the percent of missing views increases from 15\% (2 missing views) to 60\% (8 missing views), the drops of mAP and AUC increase from 0.4 and 0.2 to 7.4 and 6.9, respectively. This result shows that although the performances suffer from the noise, our aggregated feature still perform well. Compared to the noise-free image sequences, the mAP decreases slightly by 3.3 and AUC decreases by 2.2 when almost 50\% views (6 noise views) are missed. This proves that our method is still able to maintain the discrimination of aggregated features in spite of a large amount of noise.

**Background clutter.** In this experiment, we evaluate the robustness of our method against cluttered backgrounds. As Figure 5. shows, when the view sequence of a 3D shape is rendered, we place four other random shapes behind it to add background noise. The comparison can be seen in TABLE VI. and the qualitative retrieval results are shown in Figure 7. The result shows our design does improve the ability to recognize 3D shapes from cluttered background.

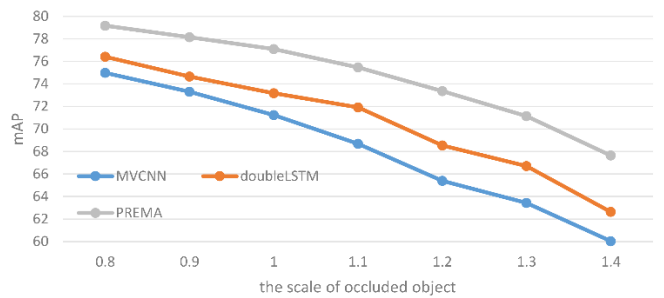

Figure 6. Performance on different scales of occluding objects, tested on ModelNet40.

**Object occlusion.** In the experiment, we place a cube around the 3D shape, causing some views to be occluded. A sample is shown in Figure 5. We compare MVCNN, doubleLSTM and PREMA with various scales of occluded objects. As the scale of the cube varies from 0.8 to 1.4, PREMA's mAP score drops from 79.17 to 67.64, doubleLSTM's drops from 76.41 to 62.63, and MVCNN's drops from 74.98 to 60.04. It turns that PREMA can take full use of residual discriminant part and has considerable robustness against object occlusion. The quantitative and qualitative results are shown in Figure 5. and Figure 5. respectively.

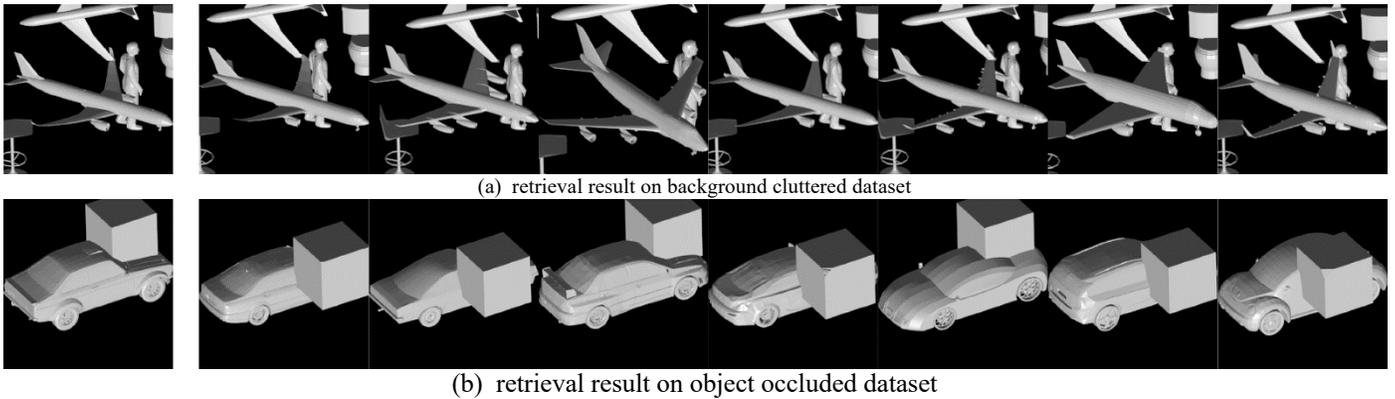

(a) retrieval result on background cluttered dataset

(b) retrieval result on object occluded dataset

Figure 7. Top 7 retrieval result on noisy data. The rightmost shape in each row is the query and the rest shapes are the top 5 retrieved shapes.

TABLE V. THE PERFORMANCE WITH AND WITHOUT REGIONAL ATTENTION UNIT.

| #missing views | AUC | mAP |
|---|---|---|
| 0 | 94.6 | 92.2 |
| 2 | 94.4 | 91.8 |
| 4 | 93.5 | 90.3 |
| 6 | 92.4 | 88.9 |
| 8 | 87.7 | 84.8 |

TABLE VI. COMPARISON UNDER THE CLUTTERED BACKGROUND ON MODELNET40 DATASET.

| Methods | clean | | clutter | |
|---|---|---|---|---|
| | AUC | mAP | AUC | mAP |
| MVCNN | 85.6 | 85.9 | 74.5 | 74.96 |
| double LSTMs | 85.6 | 85.9 | 74.5 | 74.96 |
| PREMA | **94.6** | **92.2** | **84.55** | **78.51** |

### E. Model Visualization

The goal of the proposed regional attention unit is to accentuate the coherent discriminative regions of views. And to investigate the effectiveness of it, we visualize the generated confidence maps in Figure 8. Compared with the feature maps directly output by backbone, we can find that there are some correlations among sequentially generated confidence maps: regions with high scores in different maps correspond to coherent, visually discriminant parts. When we apply the confidence maps to middle features, the activation of these parts will be enlarged, which is helpful to extract coherent sequential view features.

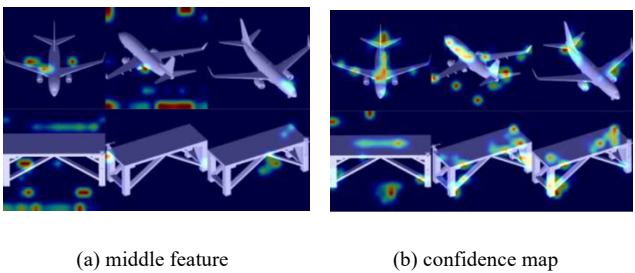

(a) middle feature     (b) confidence map

Figure 8. feature visualization. (a) The middle feature maps from the 3rd stage of ResNet. These are also input of Regional Attention Unit. (b) The confidence maps computed by Regional Attention Unit.

## V. CONCLUSION

We have presented PREMA to learn coherent discriminative 3D representation from multiple views. The model is suited for the task of 3D shape retrieval. The proposed regional attention unit takes view features and a shape feature as input and outputs a confidence map for each view, which is used to extract MCPs of the view sequence. With the ability to extract and utilize MCPs, PREMA shows great robustness to view interference, shape occlusion, and background clutter.

There are two limitations of our method. *First*, the size of the coherent discriminant part of the object is fixed in our network, making the representation not scale-invariant. *Second*, our model can only work with trained shape categories. According to each limitation, in the future, we will focus more on how to select multi-scale discriminant parts to find the scale-invariant representations. And also, cross-category generalization via combining multi-level discriminant parts is also an interesting and meaningful research task.